\documentclass[11pt]{article}

\usepackage{acl}

\usepackage{times}
\usepackage{latexsym}
\usepackage[T1]{fontenc}
\usepackage[utf8]{inputenc}
\usepackage{microtype}
\usepackage{inconsolata}
\usepackage{graphicx}
\usepackage{subcaption}
\usepackage{amsmath}
\usepackage{amssymb} 
\usepackage{graphicx}
\usepackage{multirow}
\usepackage{tabularx}
\usepackage{booktabs}
\usepackage{enumitem}
\usepackage{xspace}
\usepackage[table,svgnames]{xcolor}

\newcommand{\MethodName}{ROSE\xspace}

\title{
Reinforced Efficient Reasoning via Semantically Diverse Exploration}

\author{%
Ziqi Zhao\textsuperscript{\rm 1}\quad
Zhaochun Ren\textsuperscript{\rm 2}\quad
Jiahong Zou\textsuperscript{\rm 1}\quad
Liu Yang\textsuperscript{\rm 1}\quad
Zhiwei Xu\textsuperscript{\rm 1}\quad
Xuri Ge\textsuperscript{\rm 1}\\
\textbf{%
Zhumin Chen\textsuperscript{\rm 1}\quad
Xinyu Ma\textsuperscript{\rm 3}\quad
Daiting Shi\textsuperscript{\rm 3}\quad
Shuaiqiang Wang\textsuperscript{\rm 3}\quad
Dawei Yin\textsuperscript{\rm 3}\quad
Xin Xin\textsuperscript{\rm 1}}\thanks{Corresponding author.}\\
$^{1}$Shandong University \quad $^{2}$Leiden University \quad $^{3}$Baidu Inc.\\
\texttt{ziqizhao.work@gmail.com}  \quad  
\texttt{z.ren@liacs.leidenuniv.nl} \\
\texttt{xinxin@sdu.edu.cn} }

\begin{document}
\maketitle
\begin{abstract}

Reinforcement learning with verifiable rewards (RLVR) has proven effective in enhancing the reasoning of large language models (LLMs).
Monte Carlo Tree Search (MCTS)-based extensions improve upon vanilla RLVR (e.g., GRPO) by providing tree-based reasoning rollouts that enable fine-grained and segment-level credit assignment.
However, existing methods still suffer from limited exploration diversity and inefficient reasoning.
To address the above challenges, we propose 
\textbf{r}einforced efficient reas\textbf{o}ning via \textbf{s}emantically diverse \textbf{e}xplorations, i.e., \textbf{ROSE}, for LLMs.  
To encourage more diverse reasoning exploration, our method incorporates a semantic-entropy-based branching strategy and an $\varepsilon$-exploration mechanism. 
The former operates on already sampled reasoning rollouts to capture semantic uncertainty and select branching points with high semantic divergence to generate new successive reasoning paths, whereas the latter stochastically initiates reasoning rollouts from the root, preventing the search process from becoming overly local.
To improve efficiency, we design a length-aware segment-level advantage estimator that rewards concise and correct reasoning while penalizing unnecessarily long reasoning chains.
Extensive experiments on various mathematical reasoning benchmarks with Qwen and Llama models validate the effectiveness and efficiency of ROSE.
Codes are available at \url{https://github.com/ZiqiZhao1/ROSE-rl}.


\end{abstract}
\section{Introduction}
Reinforcement learning with verifiable rewards (RLVR) has recently been proposed to enhance the reasoning  of large language models (LLMs) in verifiable settings, including mathematical reasoning and code generation~\cite{guo2025deepseek,shao2024deepseekmath,liu2025understanding,yu2025dapo}.
Typical RLVR algorithms, such as GRPO~\cite{guo2025deepseek,shao2024deepseekmath} and its variants~\cite{yu2025dapo,liu2025understanding}, estimate the advantage of an entire rollout response based on the verified reward and uniformly propagate this advantage to all tokens within the response.

\begin{figure*}[t]
    \centering
    \begin{subfigure}[t]{0.33\textwidth}
        \centering
        \includegraphics[width=\textwidth]{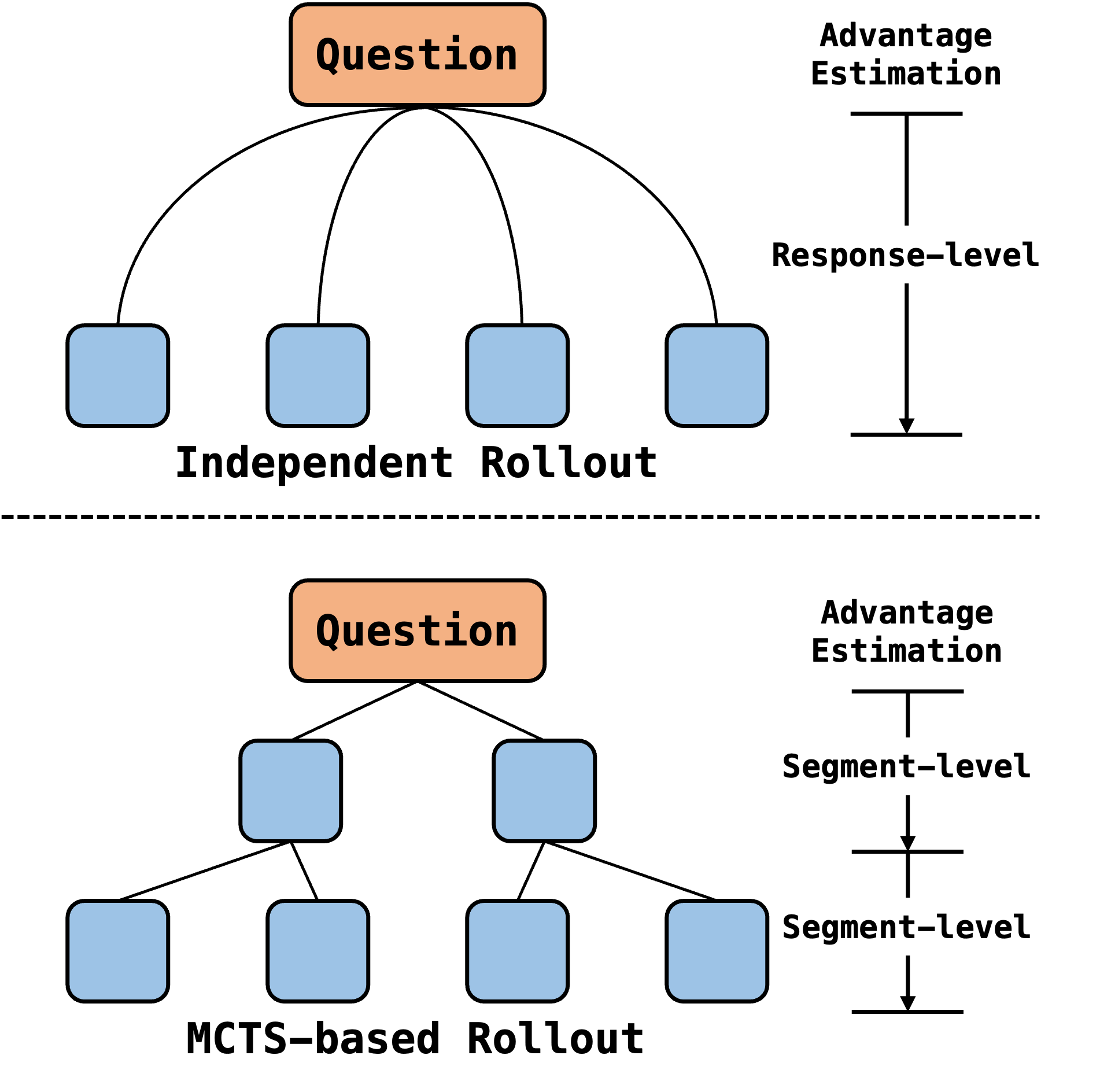} 
        \caption{}
        \label{fig:1a}
    \end{subfigure}
    \hfill
    \begin{subfigure}[t]{0.65\textwidth}  
        \centering
        \includegraphics[width=\textwidth]{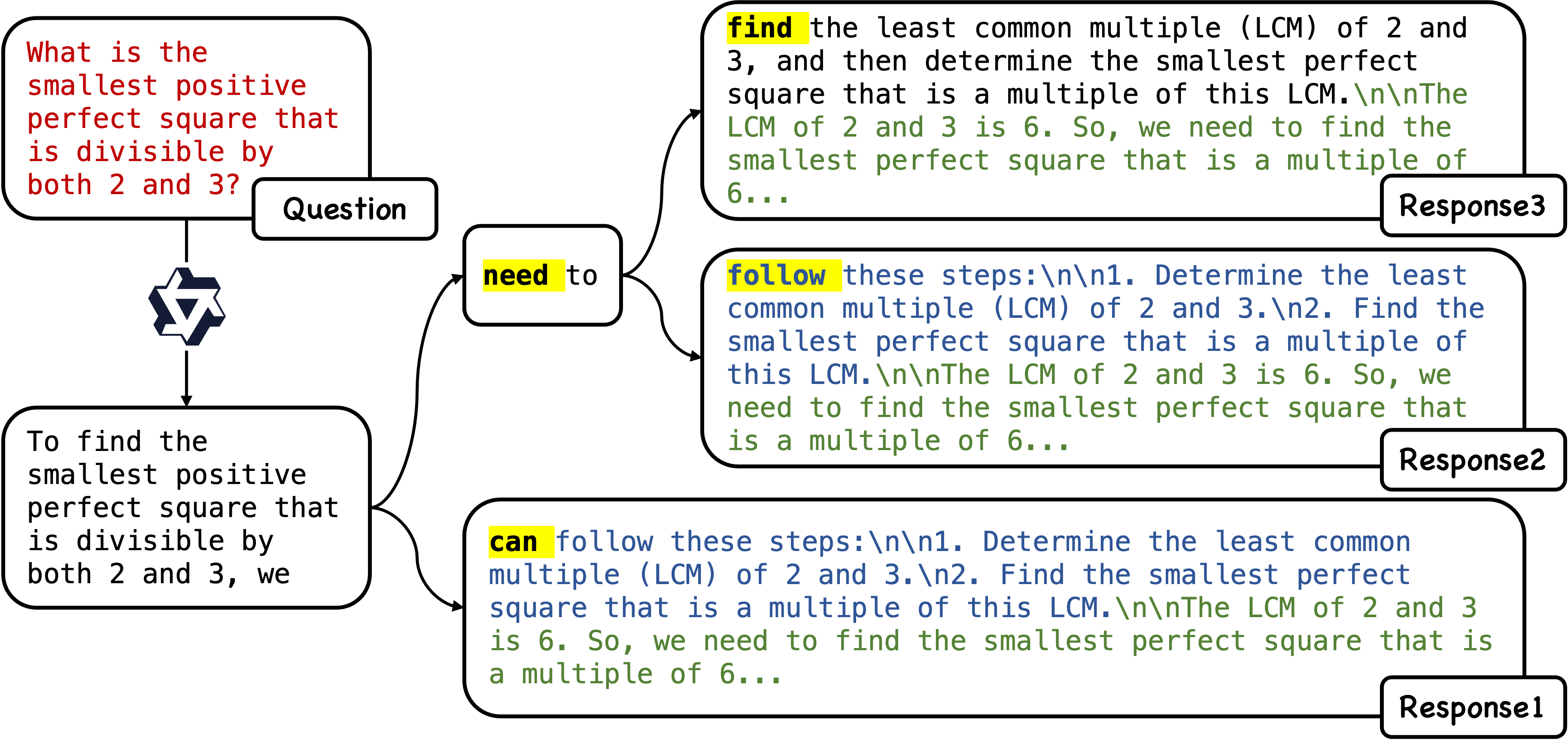}
        \caption{}
        \label{fig:1b}
    \end{subfigure}
    \caption{(a) Comparison between independent rollout (vanilla GRPO) and MCTS-based rollout. (b) Case study of generation-entropy-based branching. The tokens highlighted in yellow indicate the different tokens generated at the positions of highest entropy. Identical text across different responses is marked with the same colour (green or blue).}
    \label{fig:twocol}
\end{figure*}

While the uniform credit assignment is simple yet effective, it constrains the learning potential of the model and conflicts with human intuition.
For example, a reasoning chain that produces an incorrect response may still contain certain correct steps.
Moreover, recent studies have indicated that this training paradigm may lead to ``overthinking'', in which models are engaged in redundant reasoning~\cite{chen2024not,dai2025s,nie2026attnpo}. 
To further improve model performance, a more effective credit assignment approach is to employ Monte Carlo Tree Search (MCTS)~\cite{kocsis2006bandit} during response rollout sampling.
Unlike vanilla GRPO, which generates a group of independent responses for a given problem, MCTS enables the model to produce responses in a tree-based structure, as illustrated in Figure~\ref{fig:1a}, allowing segment-level credit assignment by computing value differences between parent and child nodes.

Despite the progress achieved by MCTS-based RLVR algorithms~\cite{li2025treepo,yang2025treerpo,zheng2025first,dong2025agentic}, \textbf{limited exploration diversity} and  \textbf{inefficient reasoning} still exist.
Specifically, most existing work uses generation entropy as the criterion for MCTS branching~\cite{zheng2025first,dong2025agentic}. 
These methods first identify the position with the highest generation entropy.
Then, tokens preceding this position are kept fixed, and successive tokens are regenerated.
Although generation entropy measures a policy's uncertainty over token selection in the current action space, this metric does not generalize well to the semantic space.
Figure~\ref{fig:1b} shows a case study of generation entropy-based branching. 
The tokens \textit{can} in response 1 and \textit{need} in response 2 correspond to the positions of highest entropy in two separate generations, yet their semantic meanings are largely the same, and the subsequent reasoning in both responses is identical (highlighted in blue).
Furthermore, although response 2 and response 3 follow different reasoning paths after branching, they are semantically similar, and their subsequent reasoning remains consistent with other responses (highlighted in green).
This indicates that current methods fail to generate semantically diverse rollouts.
Additionally, existing MCTS-based methods do not address the overthinking problem effectively, and current approaches for efficient reasoning either incur performance degradation or offer trivial performance gains~\cite{dai2025s,arora2025training}. 
How to achieve both improved performance and efficient reasoning based on MCTS remains an open question.

To address the aforementioned challenges, we propose 
\textbf{r}einforced efficient reas\textbf{o}ning via \textbf{s}emantically diverse \textbf{e}xplorations, i.e., \textbf{ROSE}, for LLMs.  
To address the first challenge, we introduce a semantic-entropy-based branching strategy together with an $\varepsilon$-exploration mechanism.
The semantic entropy metric, defined over differences in token semantics, identifies positions along a reasoning path where the model exhibits high uncertainty in the semantic space, thereby guiding the exploration toward more diverse reasoning paths.
In addition, to prevent the search process from becoming overly local, the $\varepsilon$-exploration mechanism stochastically regenerates the reasoning rollout from scratch.
Together, these methods promote more diverse and effective exploration.
To address the second challenge, we integrate credit assignment with the length of the reasoning chain. 
Leveraging the tree structure, we estimate values for each node and assign credit at the segment level.
For different correct reasoning chains originating from the same node, longer chains with deeper depth are penalized to encourage more efficient reasoning.
These components make fuller use of MCTS samples, aiming to enhance the model’s reasoning ability through more diverse and efficient exploration.
In summary, our contributions are:
\begin{itemize}
    \item We introduce a semantic-entropy guided MCTS-based rollout strategy together with an $\epsilon$-exploration mechanism, which enables more diverse exploration compared with existing approaches.
    \item We propose a segment-level advantage estimation method that incorporates reasoning length, enabling stronger performance while producing more efficient reasoning.
    \item Extensive experiments on a wide range of mathematical reasoning tasks (AIME2025, AIME2024, AMC2023, MATH500), using both Qwen and Llama models, validate the effectiveness and efficiency of our approach.
\end{itemize}
\section{Related Work}
\subsection{Reinforcement Learning for LLMs}
Reinforcement learning has been widely adopted to align LLMs with human preferences through reinforcement learning from human feedback (RLHF)~\cite{lee2024rlaif,ouyang2022training}.
More recently, reinforcement learning with verifiable rewards (RLVR) has emerged as an effective approach for enhancing the reasoning ability of LLMs~\cite{guo2025deepseek,shao2024deepseekmath,dai2025s,hao2025rethinking,zhou2026look,wu2026step,xie2025unlocking,wu2025table,an2025amo}.
By using rule-based binary (0/1) rewards to simplify reward design, GRPO~\cite{guo2025deepseek,shao2024deepseekmath} removes the need for training an extra critic model compared with vanilla PPO~\cite{schulman2017proximal}, leading to a substantial reduction in RL training overhead.
Recent studies, including DAPO~\cite {yu2025dapo}, Dr.GRPO~\cite {liu2025understanding}, VAPO~\cite{yue2025vapo}, GSPO~\cite{zheng2025group}, and CPG~\cite{chu2025gpg}, have explored improving the GRPO loss function to further enhance its reasoning capability.
In contrast to these approaches, our work focuses on improving the rollout process to enable more diverse exploration and credit assignment, without modifying the loss function. 
As a result, the proposed method is in principle compatible with a wide range of GRPO-based algorithms.

\subsection{MCTS for LLM Reasoning}
Monte Carlo Tree Search (MCTS)~\cite{kocsis2006bandit,swiechowski2023monte} offers a principled framework for exploring structured decision spaces, making it a natural candidate for performing credit assignment based on the intermediate reasoning steps.
Recent studies have explored MCTS-based sampling in RL training, showing progress on mathematical reasoning tasks~\cite{li2025treepo,yang2025treerpo,zheng2025first} as well as other complex problems~\cite{ji2025tree,dong2025agentic}.

A key challenge in applying MCTS lies in deciding where to branch, as this choice fundamentally determines the exploration trajectory and the quality of the reasoning.
Prior approaches rely on random branching~\cite{ji2025tree}, generation-entropy-based branching~\cite{dong2025agentic,zheng2025first}, branching based on fixed-length segments~\cite{li2025treepo}, or performing branching during decoding via beam search~\cite{yang2025treerpo}.
However, all these strategies fall short in promoting sufficient diverse exploration.
Meanwhile, existing methods do not explicitly account for the impact of reasoning length during advantage estimation, which can lead to overthinking during model inference.
In contrast, our approach enhances exploration diversity and enables more efficient reasoning, leading to improved performance on complex reasoning tasks.

\section{Method}
\begin{figure}[t]
    \centering
    \includegraphics[width=\linewidth]{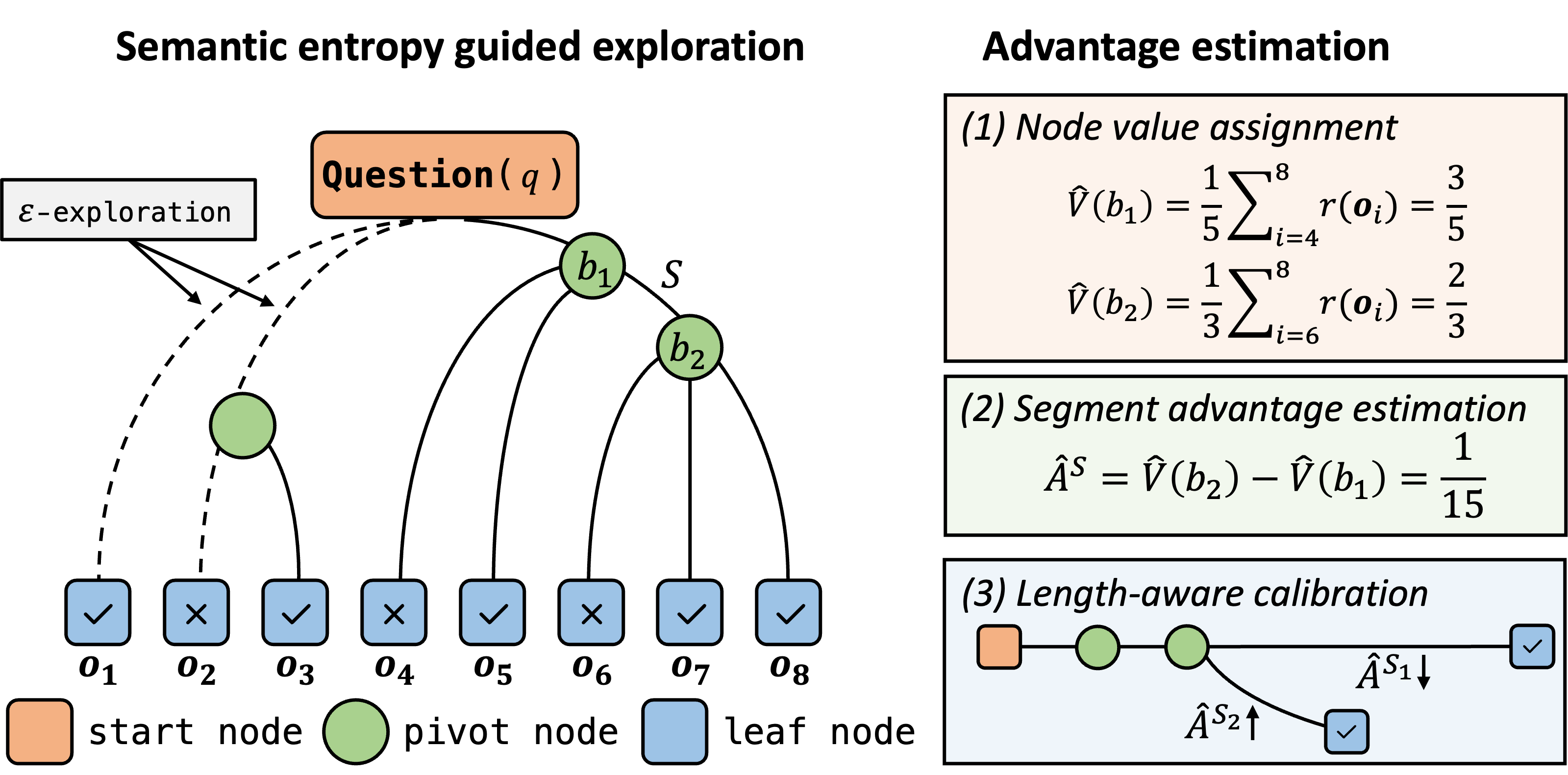}
    \caption{The overview of the \MethodName framework. The figure on the left illustrates the structure of the tree-based rollout. Pivot nodes refer to nodes with the highest semantic uncertainty, which are selected according to the semantic entropy. The rollout procedure is detailed in Section~\ref{exploration}. The figure on the right depicts the advantage estimation pipeline, comprising three stages:  (1) node value assignment, (2) segment advantage estimation and (3) length-aware calibration. These stages are described in detail in Section~\ref{advantage}.}
    \label{fig:overview}
    \vspace{-10pt}
\end{figure}

This section details \MethodName.
We first introduce how to achieve effective and diverse exploration in Section~\ref{exploration}. 
Section~\ref{advantage} then describes how tree-based exploration is leveraged to perform advantage estimation and encourage efficient reasoning. 
Finally, Section~\ref{training} presents the overall learning objective.
Figure~\ref{fig:overview} illustrates an overview of \MethodName.

\subsection{Semantic-Entropy Guided Exploration}\label{exploration}
Given a question $q$, vanilla GRPO performs rollouts by sampling a group of independent responses $\{\mathbf{o}_i\}_{i=1}^G$.
In contrast, MCTS-based methods introduce tree-structured rollouts, allowing different responses to share common prefix tokens.
The key to tree-based rollout is identifying appropriate branching positions, which encourages the model to perform more effective exploration.
A common practice is to use generation entropy to determine branching positions~\cite{dong2025agentic,zheng2025first}. 
Generation entropy provides a principled measure of the uncertainty of a policy $\pi_\theta$, i.e., the LLM,  over its action space, i.e., the vocabulary $\mathcal{V}$.
Given a question $q$ and a generated response $\mathbf{o}_i$, the generation entropy of the policy at position $k$ is defined as:
\begin{equation}\label{eq:entropy}
    \mathcal{H}_k=-\sum_{v\in\mathcal{V}}p_\theta(v|q,\mathbf{o}_{i,<k}) \log p_\theta(v|q,\mathbf{o}_{i,<k})
\end{equation}
where $p_\theta(v|q,\mathbf{o}_{i,<k})$ represents the probability distribution over the vocabulary $\mathcal{V}$ at position $k$.
Such entropy has been widely used in traditional RL~\cite{haarnoja2018soft,wang2022deep}, where different actions often exhibit significant differences, such as movement directions in a game environment~\cite{bellemare2013arcade}.
However, this assumption does not always hold in language generation.
Consider the two words \textit{can} and \textit{need} in Figure~\ref{fig:1b}. When the LLM is uncertain about which one to select, the generation entropy may be high. 
From a semantic perspective, however, this choice is actually well-determined, as both words serve the same functional role of indicating modal intent.
As a result, the responses branched from this position may exhibit extremely high similarity, and in some cases even follow identical subsequent trajectories, thereby limiting the potential for more diverse exploration.

Based on this observation, we design an additional metric to evaluate the \textbf{semantic divergence} among the current candidate tokens.
Specifically, given a question $q$ and a generated response $\mathbf{o}_i$, at position $k$, we first select the top-20 tokens from $\mathcal{V}$ with the highest probabilities to form the set $\mathcal{V}_k$ for efficiency.
Then, for each token $v_i \in \mathcal{V}_k$, its corresponding embedding $\mathbf{e}_{v_i}$ is obtained from the LLM. 
We then compute semantic divergence as the sum of pairwise similarities between all tokens in $\mathcal{V}_k$, weighted by their probabilities:

\begin{equation}\label{eq:semanticD}
\resizebox{\linewidth}{!}{$
SD_k = \displaystyle -\sum_{v_i, v_j \in \mathcal{V}_k}
    p_\theta(v_i|q,\mathbf{o}_{i,<k})
    p_\theta(v_j|q,\mathbf{o}_{i,<k})
    \cdot \cos\langle \mathbf{e}_{v_i}, \mathbf{e}_{v_j} \rangle
$}
\end{equation}
where $\cos\langle \mathbf{e}_{v_i}, \mathbf{e}_{v_j} \rangle$ represents the cosine similarity between the embeddings of tokens $v_i$ and $v_j$.
The key idea of semantic divergence is that when the high-probability tokens exhibit large semantic differences, the current position becomes an ideal branching point, leading to more distinct subsequent reasoning paths.

Finally, we define \textbf{semantic entropy}  as the product of generation entropy and semantic divergence, and use it as the branching indicator:
\begin{equation}\label{eq:semanticE}
SE_k=SD_k \cdot \mathcal{H}_k
\end{equation}
This combined measure captures both probabilistic uncertainty and semantic dispersion, allowing ROSE to more accurately identify positions where alternative continuations are more likely to lead to genuinely diverse reasoning paths.

The rollout process based on branching metrics is summarized as follows. 
Given a question $q$, a complete response is first generated. 
For each position in the generated response, the proposed branching metric is computed, and the position with the highest value is selected. 
Then a new response is regenerated at this position, keeping the preceding part of the response unchanged. 
The corresponding metric of the newly generated sequence is computed, and the selection is then performed on all existing rollout responses. 
The whole process is repeated until the number of generated responses reaches the predefined parameter $G$.

In addition, inspired by the $\varepsilon$-greedy~\cite{sutton1998reinforcement} strategy in reinforcement learning, we propose an \textbf{$\varepsilon$-exploration} mechanism. 
Specifically, before generating each response, there is an $\varepsilon$ probability of generating the response from scratch, i.e., rolling out an independent response; otherwise, the rollout follows the proposed semantic-entropy-based branching strategy.
This mechanism prevents the search from becoming overly focused on local regions and further balances the depth and breadth of exploration.
After completing the rollout process for a given query, we can obtain a tree structure, an example of which is shown on the left side of Figure~\ref{fig:overview}.
During rollout, we apply dynamic sampling~\cite{yu2025dapo} to remove groups whose responses receive identical rewards, improving efficiency.
The proposed methods offer an effective exploration-exploitation tradeoff to better search the reasoning paths for LLMs.

\subsection{Advantage Estimation}\label{advantage}
Based on the tree-structured exploration, we perform segment-level credit assignment through (1) node value assignment, (2) segment advantage estimation and (3) length-aware calibration.

\noindent\textbf{Node value assignment.}
After completing tree-structured sampling, a single response may contain multiple branching nodes.
Including the start and the terminal positions, these nodes partition a response into several consecutive segments.
Formally, for a response $\mathbf{o}_i$ to a given question $q$, let ${b_0, b_1, \dots, b_k}$ denote the node positions, where $b_0$ is the start position, $b_1, \dots, b_{k-1}$ correspond to the pivot positions immediately before each branching point, and $b_k$ is the leaf (terminal) position. The response can then be decomposed as
\begin{equation}
\mathbf{o}_i=\bigcup_{j=1}^k \mathbf{o}_{i,b_{j-1} < t \leq b_j}, \text{with } b_0=0, b_k=|\mathbf{o}_i|
\end{equation}
Under this partition, each segment is initiated at either the start position or a branching position selected based on maximal semantic entropy observed during the rollout stage.
For a pivot node $b_j$ with $0<j<k$, we define the set of responses that contain this node as:
\begin{equation}
    \Omega_{b_j} = \{\mathbf{o}_m|\mathbf{o}_m\text{ traverses the pivot node } b_j\}
\end{equation}
For the start node $b_0$, we define $\Omega_{b_0}$ as the set of all responses, i.e., $\Omega_{b_0}=\{\mathbf{o}_m\}_{m=1}^G$. For the leaf node, we define $\Omega_{b_k}$ as the set of $\mathbf{o}_i$, i.e., $\Omega_{b_k}=\{\mathbf{o}_i\}$.

Next, we define the value of  node $b_j$ with $0\leq j\leq k$ as the average reward of responses in $\Omega_{b_j}$.
\begin{equation}
\hat{V}(b_j)=\frac{1}{|\Omega_{b_j}|}\sum_{\mathbf{o}_m \in \Omega_{b_j}}r(\mathbf{o}_m)
\end{equation}
where $r()$ denotes a rule-based reward function that evaluates the correctness of each response and assigns a binary reward (1 for correct and 0 for incorrect).

\noindent\textbf{Segment advantage estimation.}
Next, we can compute the segment-level advantage between two nodes based on the values assigned to the nodes. 
According to the definition of node value, the value of a node can be interpreted as the probability of deriving a correct reasoning chain starting from that node.
Therefore, the reasoning contribution of the segment is quantified by the difference between the two node values.
Specifically, for any token $\mathbf{o}_{i,t} \in \mathbf{o}_i$ with $b_{j-1} < t \leq b_j$, the advantage of $\mathbf{o}_{i,t} $ is defined as:
\begin{equation}
    \hat{A}_{i,t}=\hat{V}(b_j)-\hat{V}(b_{j-1})
\end{equation}

\noindent\textbf{Length-aware calibration.}
Furthermore, although multiple reasoning paths may lead to correct outcomes, we aim to encourage the model to adopt more efficient reasoning and avoid overthinking. 
To this end, we apply a length-aware calibration to the advantages of responses that are correct but require an excessive number of tokens.
Specifically, we first identify the shortest correct response $\mathbf{o}_s$.
Then, for every other correct response $\mathbf{o}_c$, we locate the pivot node $b_c$ at which $\mathbf{o}_s$ and $\mathbf{o}_c$ diverge. 
That is, prior to $b_c$, the two responses share an identical subsequence $\mathbf{o}_{s,\le b_c} = \mathbf{o}_{c,\le b_c}$, whereas after $b_c$ they follow distinct continuations $\mathbf{o}_{s,> b_c}$ and $\mathbf{o}_{c,> b_c}$, respectively.
A length-proportional calibration is then applied to the longer response, thereby encouraging the model to produce more efficient reasoning. 
Specifically, for each token $\mathbf{o}_{c,t} \in \mathbf{o}_{c}$ with $t>b_c$, its advantage is updated according to the following rule:
\begin{equation}
    \hat{A}_{i,t} \gets \hat{A}_{i,t}-|\hat{A}_{i,t}|\cdot (1-(\frac{|\mathbf{o}_s|-b_c}{|\mathbf{o}_c|-b_c})^\alpha)
\end{equation}
where $\alpha$ is a hyperparameter controlling the extent of the adjustment.
The ratio $\frac{|\mathbf{o}_s|-b_c}{|\mathbf{o}_c|-b_c}$ measures the relative lengths of the two reasoning branches after their divergence at $b_c$. 
Since the two responses share an identical reasoning prefix before $b_c$, their post-divergence segments can be directly compared: the more efficient branch receives a higher advantage, while the longer branch incurs a length-proportional adjustment.

\subsection{Model Training}\label{training}
We adopt the improved modifications of vanilla GRPO's optimizing objective proposed in Dr.GRPO~\cite{liu2025understanding} as the training objective, together with a KL penalty term:


\begin{equation}
\label{eq:grpoloss}
\resizebox{0.95\linewidth}{!}{$
\begin{aligned}
\mathcal{L}_\text{\MethodName}(\theta) 
&=  -\frac{1}{G}\sum_{i=1}^{G}\sum_{t=1}^{|\mathbf{o}_i|} \Bigg( 
\min \Big( r_{i,t}(\theta) \hat{A}_{i,t}, 
[r_{i,t}(\theta)]_{1-\epsilon'}^{1+\epsilon'}\hat{A}_{i,t} \Big) \\
\hspace{0pt}
&
- \beta D_{\text{KL}}(\pi_{\theta} \| \pi_{\text{ref}}) 
\Bigg),
\end{aligned}
$}
\end{equation}
where
\begin{equation}
    r_{i,t}(\theta)=\frac{\pi_{\theta}(\mathbf{o}_{i,t} \mid q, \mathbf{o}_{i,<t})}{\pi_{\theta_{\text{old}}}(\mathbf{o}_{i,t} \mid q,\mathbf{o}_{i,<t})},
\end{equation}
$\pi_{\text{old}}$ denotes sampling model, $\pi_{\text{ref}}$ denotes reference model and operator $[r_{i,t}(\theta)]_{1-\epsilon'}^{1+\epsilon'}$ clips the ratio to $[1-\epsilon',1+\epsilon']$.

\section{Experimental Setup}

\subsection{Datasets and Metrics}
For the training dataset, following prior studies~\cite{zhu2025surprising, liu2025understanding}, we use MATH~\cite{hendrycksmath2021}, which contains 7,500 problems.
For evaluation, four publicly available standard mathematical reasoning benchmarks are considered, including 
AIME2024, AIME2025, AMC23,  
and MATH500.
MATH500 is a subset of the test split of the MATH dataset, consisting of 500 problems.
During validation, we sample 8 responses for each question and adopt pass@8 as the primary metric for assessing the performance of model reasoning.
pass@$k$ measures whether at least one of the $k$ sampled responses correctly solves a given problem. 
Unlike prior work that relies on the mean@$k$ metric~\cite{li2025treepo,yang2025treerpo}, which reflects the average accuracy across all samples, pass@$k$ more effectively captures the model’s ability to solve previously challenging problems that it might fail to answer.

\begin{table*}[t]
\centering
\resizebox{0.93\textwidth}{!}{
\renewcommand{\arraystretch}{1}

\begin{tabular}{
 ll
cccccc
c 
}
\toprule
\multirow{2}{*}{\textbf{Model}} 
& \multirow{2}{*}{\textbf{Dataset}}
& \multirow{2}{*}{\textbf{Base Model}}
& \multicolumn{3}{c}{\textbf{GRPO Variants}}
& \multicolumn{2}{c}{\textbf{MCTS-Based}}
& \multirow{2}{*}{\textbf{\MethodName}} \\
\cmidrule(lr){4-6}
\cmidrule(lr){7-8}

& 
& 
& \small{GRPO}
& \small{DAPO}
& \small{Dr.GRPO}
& \small{TreePO}
& \small{FR3E}
& \\
\midrule

\multirow{5}{*}{\texttt{Qwen3-4B-Base}}
& AIME2024 & 13.33 & 16.67 & 16.67 & 16.67 & 16.67 & 16.67 & \textbf{23.33}{\small\color{red}{+6.67}} \\
& AIME2025 & 16.67 & 20.00 & 16.67 & 23.33 & 13.33 & 20.00 & \textbf{23.33}{\small\color{red}{+3.33}} \\
& MATH500  & 74.00 & 79.8 & 79.00 & 78.60 & \textbf{82.00} & 80.00 & 80.80{\small\color{blue}{-1.20}} \\
& AMC23    & 45.00 & \textbf{77.50} & 75.00 & 70.00 & 72.50 & 75.00 & \textbf{77.50}{\small\color{red}{+0.00}} \\
\cmidrule(lr){2-9}
& Average  & 37.25 & 48.49 & 46.83 & 47.14 & 46.12 & 47.92 & \textbf{51.24}{\small\color{red}{+2.75}} \\
\midrule

\multirow{5}{*}{\texttt{Qwen3-8B-Base}}
& AIME2024\space\space\space & 13.33 & 23.33 & 26.67 & 26.67 & 23.33 & 23.33 & \textbf{33.33}{\small\color{red}{+6.67}} \\
& AIME2025 & 10.00 & 23.33 & 23.33 & 23.33 & 23.33 & 23.33 & \textbf{30.00}{\small\color{red}{+6.67}} \\
& MATH500  & 68.20 & 79.40 & 79.40 & 81.60 & \textbf{84.20} & 80.80 & 83.00{\small\color{blue}{-1.20}} \\
& AMC23    & 47.50 & 72.50 & 75.00 & 72.50 & 70.00 & 75.00 & \textbf{80.00}{\small\color{red}{+5.00}} \\
\cmidrule(lr){2-9}
& Average  & 34.76 & 49.64 & 51.10 & 51.02 & 50.21 & 50.62 & \textbf{55.75}{\small\color{red}{+4.65}} \\
\midrule

\multirow{5}{*}{\texttt{Llama-3.2-3B-Ins.}}
& AIME2024 & 10.00 & {16.67} & {16.67} & 13.33 & {16.67} & {16.67} & \textbf{20.00}{\small\color{red}{+3.33}} \\
& AIME2025 & 0.00  & 3.33  & 3.33  & \textbf{6.67}  & 3.33  & \textbf{6.67}  & \textbf{6.67}{\small\color{red}{+0.00}} \\
& MATH500  & 46.00 & 53.40 & 54.60 & 54.40 & 52.60 & 54.40 & \textbf{55.00}{\small\color{red}{+0.40}} \\
& AMC23    & 35.00 & 40.00 & 37.50 & 40.00 & 35.00 & 37.50 & \textbf{45.00}{\small\color{red}{+5.00}} \\
\cmidrule(lr){2-9}
& Average  & 22.75 & 28.35 & 28.02 & 28.60 & 26.90 & 28.81 & \textbf{31.67}{\small\color{red}{+2.86}} \\
\bottomrule
\end{tabular}
}
\caption{Experimental results with pass@8 metric (\%). For each test dataset, we report the best scores achieved during training. Boldface denotes the best results under each dataset. The absolute improvement or degradation compared to the second-best score is also indicated.}
\label{tab:main_results}
\vspace{-15pt}
\end{table*}


\subsection{Model and Baselines}
To provide a more comprehensive comparison of the proposed method, we evaluate it using backbone models from two model families, Qwen and Llama, with different parameter scales, including \texttt{Llama-3.2-3B-Instruct}~\cite{grattafiori2024llama}, \texttt{Qwen3-4B-Base}, and \texttt{Qwen3-8B-Base}~\cite{yang2025qwen3}.
The \texttt{Qwen3} models have two modes (thinking and non-thinking), and the non-thinking mode is adopted for both training and inference.

Comparisons are performed between \MethodName and existing approaches, which mainly fall into two categories: GRPO-based variants and MCTS-based methods.
The GRPO-based variants include vanilla GRPO~\cite{guo2025deepseek,shao2024deepseekmath}, Dr.GRPO~\cite{liu2025understanding}, and DAPO~\cite{yu2025dapo}. 
Dr.GRPO computes advantages as deviations from the group mean, without variance normalization, and removes the length normalization term from the loss function.  
DAPO improves GRPO by incorporating techniques such as clip-higher and rejection sampling.

The MCTS-based baselines include FR3E~\cite{zheng2025first} and TreePO~\cite{li2025treepo}. 
FR3E is a representative MCTS-based method that determines branching positions based on generation entropy and adopts a two-step framework for segment-level advantage computation.
Besides, TreePO structures the rollout process as a tree by branching at fixed-length segments and computes advantages over the resulting sub-trees. 

\subsection{Implementation Details}
All experiments are conducted using the VeRL framework~\cite{sheng2025hybridflow} in this paper.
For RL training, we set the batch size to 512, the number of rollouts per prompt as $G=8$, the learning rate to $1\times10^{-6}$, the clipping ratio as $\epsilon'=0.2$, the KL divergence coefficient as $\beta=0.001$, and the maximum number of training epochs to 8.
For evaluation, the temperature is set to 0.6, top-$p$ sampling is applied with $p=0.95$, and 8 candidate responses are sampled per prompt.
Prompts whose lengths exceed 2048 tokens are filtered out, and the maximum generation length is set to 4096 tokens.
The probability of generating the response from scratch $\varepsilon$ is set to 0.5 by default, and the coefficient for length-aware calibration $\alpha$ is searched from $\{0.5,1,2,3\}$.
Our experiments are conducted on 8 $\times$ NVIDIA A800 (80G) GPUs.

\section{Experimental Results}
\subsection{Overall Performance}

\begin{figure*}[t]
  \includegraphics[width=1\linewidth]{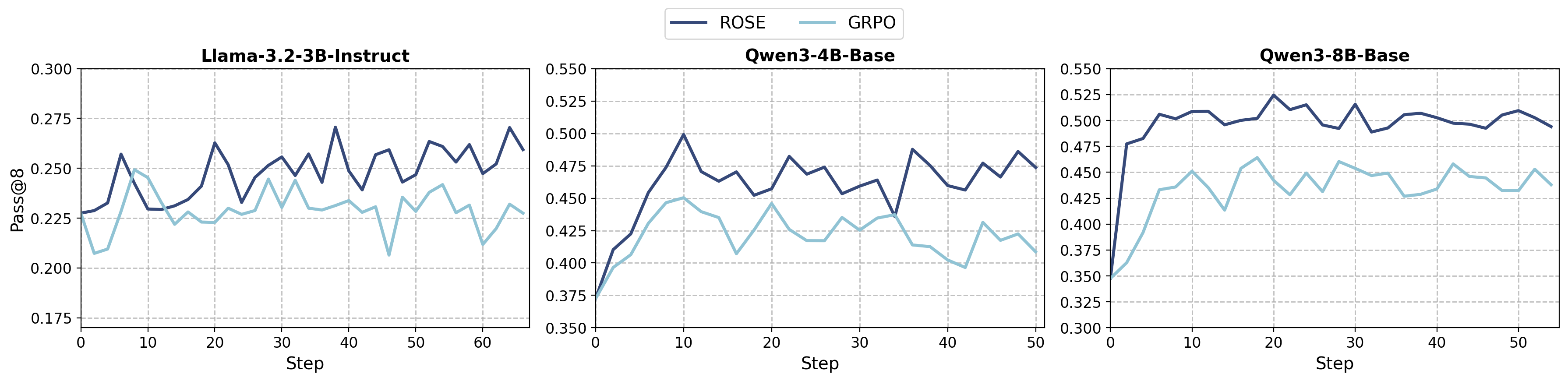}
  \caption {Learning curves. Average performance across four datasets as training progresses.}
  \label{fig:learning_curve}
  \vspace{-10pt}
\end{figure*}

\noindent\textbf{Accuracy evaluation.}
Table~\ref{tab:main_results} presents the experimental results of all methods. 
It can be observed that \MethodName achieves substantial improvements over the strongest baseline in most settings.
DAPO and Dr.GRPO are variants that modify the GRPO loss function.
However, they do not yield consistent or substantial improvements, with performance gains observed only in certain scenarios, such as models with larger parameter scales.
Among MCTS-based approaches, TreePO and FR3E achieve performance comparable to GRPO and its variants. 
In particular, TreePO yields pronounced improvements on the in-domain dataset MATH500 but performs worse on other benchmarks.
This suggests that its fixed-length branching strategy fails to induce more diverse reasoning trajectories during exploration, limiting out-of-domain generalization.

For \MethodName, significant performance gains are first observed on more challenging tasks, indicating that the method facilitates more divergent exploration during the rollout phase, which is beneficial for solving difficult problems.
In addition, \MethodName consistently yields notable improvements across different model scales. 
Larger models typically encapsulate richer knowledge, and the proposed approach appears to leverage this capacity more effectively, resulting in greater performance gains.
Finally, recent studies have suggested that models in the Qwen family may suffer from potential data leakage~\cite{wu2025reasoning}. 
Nevertheless, comparable performance gains are also observed on Llama models of similar parameter scales, indicating that the improvements are not confined to a specific model family.

\begin{table}[t]
\centering
\setlength{\tabcolsep}{2pt}
\resizebox{1\linewidth}{!}{
\renewcommand{\arraystretch}{1.1}
\begin{tabular}{l|ccccc}
\toprule
\textbf{\small Metric} & \textbf{\small AIME2024} & \textbf{\small AIME2025} & \textbf{\small MATH500} & \textbf{\small AMC23} & \textbf{\small Average} \\
\toprule
\rowcolor{Gainsboro} \multicolumn{6}{l}{\texttt{Llama-3.2-3B-Instruct}} \\
Generation Entropy                & 16.67 & \textbf{6.67} & 55.2 & 42.5 & 30.26 \\
Semantic Divergence    & \textbf{20.00} & \textbf{6.67} & 54.6 & 42.5 & 30.94 \\
Semantic Entropy       & \textbf{20.00} & \textbf{6.67} & \textbf{55.0} & \textbf{45.00} & \textbf{31.67} \\
\midrule

\rowcolor{Gainsboro} \multicolumn{6}{l}{\texttt{Qwen3-4B-Base}} \\
Generation Entropy                 & 20.00 & 23.33 & 79.80 & 72.5 & 48.91 \\
Semantic Divergence    & 20.00 & \textbf{26.67} & \textbf{80.80} & 75.00 & 50.62 \\
Semantic Entropy       & \textbf{23.33} & 23.33 & \textbf{80.80} & \textbf{77.50} & \textbf{51.24} \\
\midrule

\rowcolor{Gainsboro} \multicolumn{6}{l}{\texttt{Qwen3-8B-Base}} \\
Generation Entropy                 & 20.00 & 23.33 & 81.20 & 72.5 & 49.26 \\
Semantic Divergence    & 30.00 & {26.67} & \textbf{83.00} & 75.00 & 53.67 \\
Semantic Entropy       & \textbf{33.33} & \textbf{30.00} & \textbf{83.00} & \textbf{80.00} & \textbf{55.75} \\
\bottomrule
\end{tabular}
}
\caption{Experimental results with pass@8 metric (\%) for different branching metrics.}
\vspace{-15pt}
\label{tab:metric}
\end{table}
\begin{figure}[t]
  \includegraphics[width=1\linewidth]{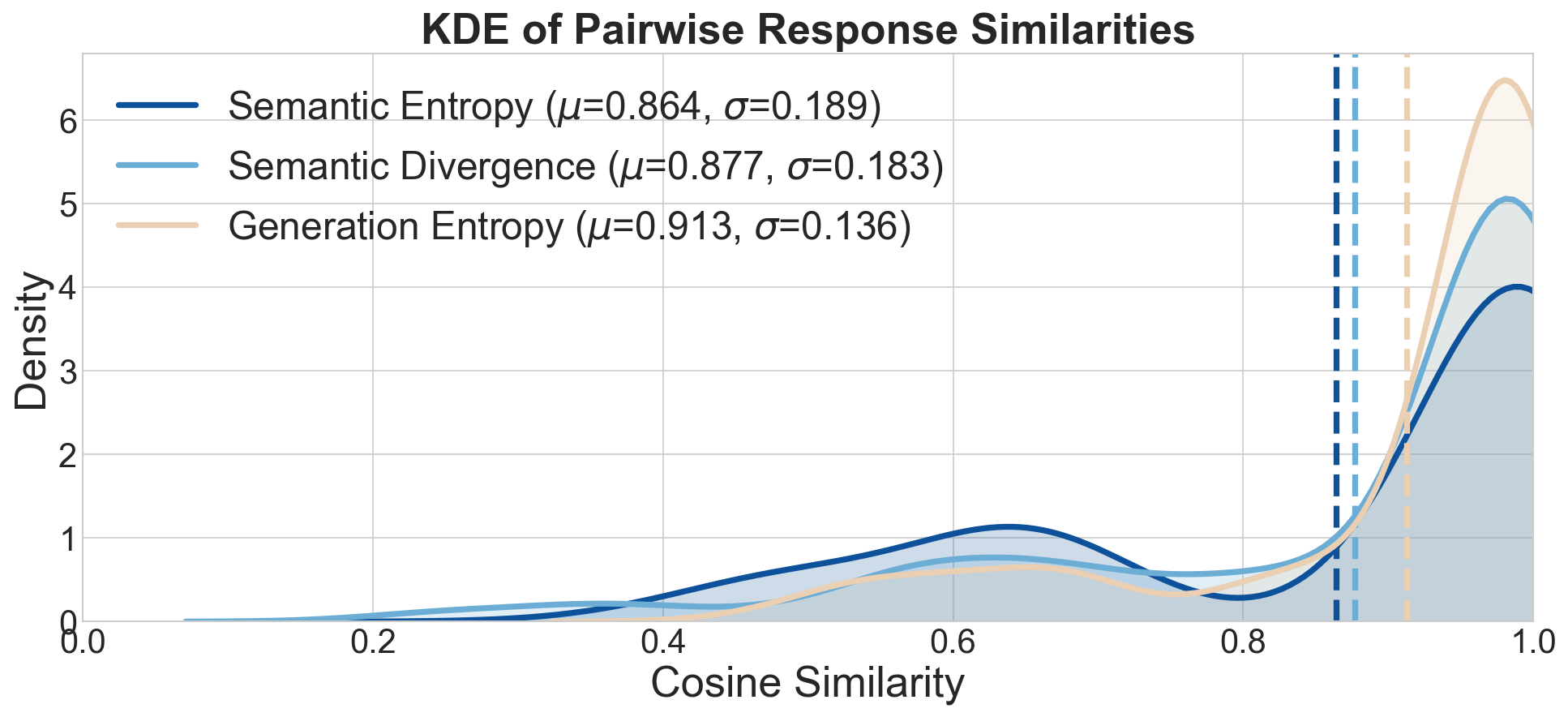}
  \caption {Kernel density estimation (KDE) of pairwise sentence similarities. The dashed line indicates the average cosine similarity. }
  \label{fig:kde}
  \vspace{-10pt}
\end{figure}

\noindent\textbf{Learning dynamics.}
Figure~\ref{fig:learning_curve} presents the learning curves of GRPO and \MethodName. 
Across different model scales, \MethodName exhibits a clear performance improvement over the vanilla GRPO. 
Moreover, as the model scale increases, the learning curves of \MethodName become noticeably more stable.
After convergence, \MethodName maintains stable and competitive performance, whereas the vanilla GRPO shows noticeable fluctuations and fails to achieve significant improvements in small-scale (\texttt{3B} and \texttt{4B}) models.

\subsection{Branching Metric Analysis}

\noindent\textbf{Performance comparison.}
Table~\ref{tab:metric} presents the experimental results for three different branching metrics (entropy, semantic divergence, and semantic entropy). 
For all branching metrics, the probability $\varepsilon$ is fixed to 0.5.
The results show that entropy achieves performance comparable to vanilla GRPO, indicating that it fails to effectively distinguish uncertain regions in the model’s reasoning trajectories. 
In contrast, both semantic divergence and semantic entropy yield consistent improvements over entropy, with semantic entropy exhibiting greater robustness across different datasets.

\noindent\textbf{Analysis of reasoning diversity.} 
To further analyze the differences among branching metrics, we conduct a quantitative analysis of the diversity of responses generated during the rollout phase under different branching metrics. 
Specifically, for a fixed batch of questions, rollouts are performed using different metrics, and the pairwise cosine similarity between embeddings of multiple responses corresponding to the same question is computed. 
The embeddings are obtained using the \texttt{Qwen-text-embedding-v4} model. 
The resulting similarity distributions are visualized using kernel density estimation (KDE), as shown in Figure~\ref{fig:kde}.

As illustrated in the figure, the distributions induced by our methods exhibit lower peaks and heavier tails. 
The mean similarities of semantic entropy and semantic divergence are comparable and both are lower than those of entropy, indicating a higher degree of dispersion among generated responses.
Such increased reasoning diversity encourages broader exploration of the solution space, which aligns with the observed performance gains on more challenging benchmarks.

\begin{table}[t]
\centering
\setlength{\tabcolsep}{2pt}
\resizebox{1\linewidth}{!}{
\renewcommand{\arraystretch}{1.1}
\begin{tabular}{l|cc|cc}
\toprule
\multirow{2}{*}{\textbf{Method}} 
& \multicolumn{2}{c|}{{\texttt{Llama-3.2-3B-Instruct}}} 
& \multicolumn{2}{c}{{\texttt{Qwen3-8B-Base}}} \\
\cline{2-5}
& \textbf{Pass@8} & \textbf{Length} & \textbf{Pass@8} & \textbf{Length} \\
\midrule
Base Model 
& 22.75 {\small\color{red}{+0.0}} 
& 788.6 {\small\color{red}{+0.0\%}} 
& 34.76 {\small\color{red}{+0.0}} 
& 932.7 {\small\color{red}{+0.0\%}} \\

GRPO 
& 28.35 {\small\color{red}{+5.6}} 
& 804.8 {\small\color{blue}{+2.1\%}} 
& 49.64 {\small\color{red}{+14.9}} 
& 907.3 {\small\color{red}{-2.7\%}} \\

Dr.GRPO 
& 28.60 {\small\color{red}{+5.9}} 
& 794.5 {\small\color{blue}{+0.7\%}} 
& 51.02 {\small\color{red}{+16.3}} 
& 930.6 {\small\color{red}{-0.2\%}} \\

\rowcolor{Gainsboro} \multicolumn{5}{l}{\MethodName} \\

 {\space\space $\vdash\alpha=0$} 
& 30.98 {\small\color{red}{+8.2}} 
& 733.1 {\small\color{red}{-7.0\%}} 
& 55.17 {\small\color{red}{+20.4}} 
& 907.2 {\small\color{red}{-2.7\%}} \\

 {\space\space $\vdash\alpha=1$} 
& \textbf{31.67} {\small\color{red}{+8.9}} 
& 702.0 {\small\color{red}{-11.0\%}} 
& \textbf{55.75} {\small\color{red}{+21.0}} 
& 904.4 {\small\color{red}{-3.0\%}} \\

 {\space\space $\vdash\alpha=2$} 
& 31.29 {\small\color{red}{+8.5}} 
& 715.4 {\small\color{red}{-9.3\%}} 
& 54.87 {\small\color{red}{+20.1}} 
& 897.2 {\small\color{red}{-3.8\%}} \\

 {\space\space $\vdash\alpha=3$} 
& 30.78 {\small\color{red}{+8.0}} 
& 692.6 {\small\color{red}{-12.2\%}} 
& 55.12 {\small\color{red}{+20.4}} 
& 885.8 {\small\color{red}{-5.0\%}} \\

 {\space\space $\vdash\alpha=10$} 
& 29.06 {\small\color{red}{+6.3}} 
& 634.3 {\small\color{red}{-19.6\%}} 
& 54.29 {\small\color{red}{+19.5}} 
& 860.4 {\small\color{red}{-7.8\%}} \\
\bottomrule
\end{tabular}
}
\caption{Experimental results with pass@8 (\%) and length (token counts) metrics. }
\vspace{-15pt}
\label{tab:length}
\end{table}

\subsection{Reasoning Efficiency Analysis}
To investigate whether \MethodName can achieve more efficient reasoning, we evaluate different values of the hyperparameter $\alpha$ and report the corresponding pass@8 scores and reasoning lengths. 
The averaged results across the four datasets are presented in Table~\ref{tab:length}.

The results show a clear trend that increasing $\alpha$ reduces the reasoning length while maintaining strong pass@8 performance. 
In particular, moderate values of $\alpha$ (e.g., $\alpha=1$ or $\alpha=2$) yield the best trade-off between accuracy and efficiency, achieving higher pass@8 scores together with substantial reductions in reasoning length. 
Even with a relatively large value of $\alpha$ (i.e., $\alpha=10$), our method still consistently outperforms GRPO variants in terms of pass@8.
Overall, these results demonstrate that \MethodName enables more efficient reasoning without sacrificing task performance.
The evolution of response length across training steps is presented in Appendix~\ref{sec:length_dynamics}.

\begin{table}[t]
\centering
\setlength{\tabcolsep}{2pt}
\resizebox{1\linewidth}{!}{
\renewcommand{\arraystretch}{1.1}
\begin{tabular}{l|ccc}
\toprule
\textbf{Method} & \texttt{L-3B-Ins.} & \texttt{Q-4B-Base} & \texttt{Q-8B-Base} \\
\midrule
\MethodName & \textbf{31.67} & \textbf{51.24} & \textbf{55.75} \\
\midrule
$\vdash$ w/o $\varepsilon$-exploration & 26.44 & 48.81 & 49.27 \\
$\vdash$ w/ random branching & 29.94 & 48.05 & 49.02 \\
$\vdash$ w/o advantage estimation& 30.43 & 49.21 & 52.32 \\
\bottomrule
\end{tabular}
}
\vspace{-5pt}
\caption{Experimental results with pass@8 metric (\%). \texttt{L} denotes \texttt{Llama-3.2} and \texttt{Q} denotes \texttt{Qwen3}.}
\label{tab:ablation}
\vspace{-15pt}
\end{table}

\subsection{Ablation Study}
An ablation study is conducted to analyze the contribution of each component, with the results presented in Table~\ref{tab:ablation}.
\textbf{(1) w/o $\varepsilon$-exploration} removes the $\varepsilon$-possibility branching mechanism (i.e., $\varepsilon$ = 0), which results in consistent performance drops across all backbone models, indicating that the model tends to fall into overly local exploration and loses exploration diversity.
Additional results examining different values of $\varepsilon$ are provided in Appendix~\ref{sec:varepsilon}.
\textbf{(2) w/ random branching} randomly determines the branching positions during rollout, which leads to performance degradation, indicating that the semantic entropy metric can effectively identify uncertain points along the reasoning trajectories.
\textbf{(3) w/o advantage estimation} removes the segment-level advantage estimation and instead directly uses the GRPO advantage formulation and loss function. 
This modification leads to degraded performance, which suggests that segment-level advantage estimation plays an important role in shaping learning signals during reasoning.

\subsection{Case Study}
We also conduct case studies and observe that our semantic-entropy-based approach can identify semantical uncertain positions along the reasoning paths,  encouraging more diverse reasoning. 
Detailed examples are provided in the Appendix~\ref{sec:case_study}.

\subsection{Response Length Dynamics}\label{sec:length_dynamics}

\begin{figure*}[t]
\centering
  \includegraphics[width=1\linewidth]{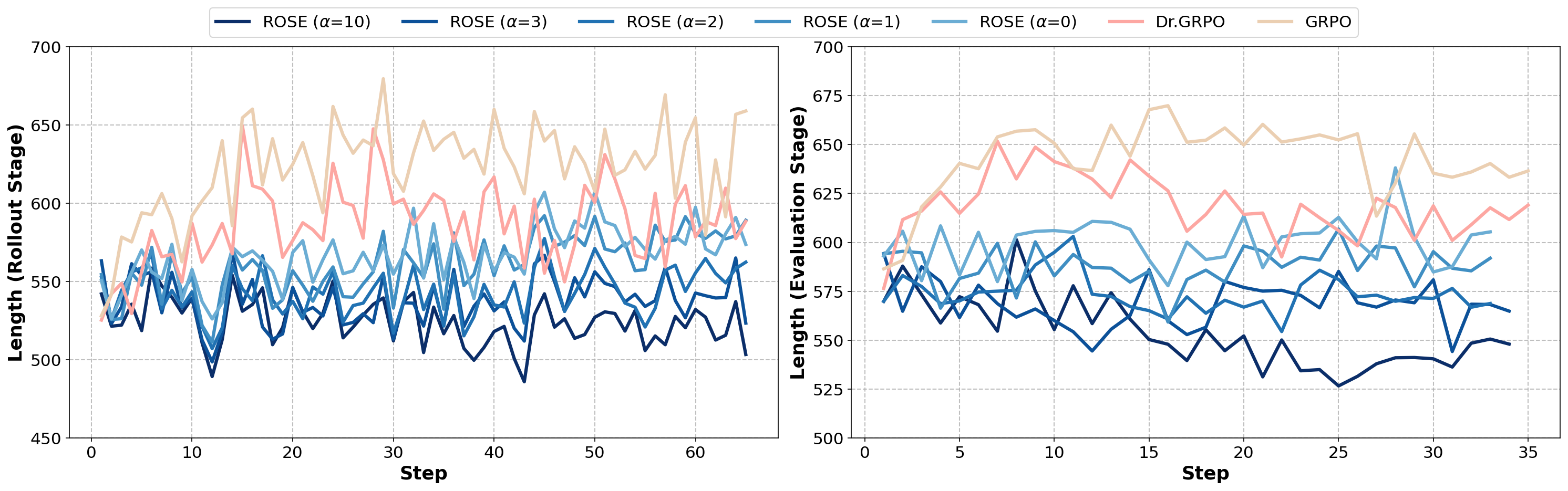}
  \caption {The average response length per prompt during the rollout stage (left) and the evaluation stage (right)}
  \label{fig:length}
\end{figure*}

To further investigate the effect of the hyperparameter $\alpha$, we examine how the average response length per prompt during the rollout stage and the evaluation stage evolves over training steps, with the results presented in Figure~\ref{fig:length}. 
We observe that as $\alpha$ increases, the generated response lengths in both the rollout and evaluation stages decrease substantially, and are consistently shorter than those of Dr.GRPO and GRPO. 
This indicates that our method can effectively regulate the generation length by adjusting $\alpha$, thereby enabling more efficient reasoning.

\begin{table}[t]
\centering
\setlength{\tabcolsep}{2pt}
\resizebox{1\linewidth}{!}{
\renewcommand{\arraystretch}{1.1}
\begin{tabular}{l|ccccc}
\toprule
\textbf{\small Metric} & \textbf{\small AIME2024} & \textbf{\small AIME2025} & \textbf{\small MATH500} & \textbf{\small AMC} & \textbf{\small Average} \\
\toprule

\rowcolor{Gainsboro} \multicolumn{6}{l}{\texttt{Llama-3.2-3B-Instruct}} \\
$\varepsilon {=} 0$      & 13.33 & 3.33 & 51.60 & 37.50 & 26.44 \\
$\varepsilon {=} 0.3$    & \textbf{20.00} & 3.33 & 54.60 & 37.50 & 28.86 \\
{$\varepsilon {=} 0.5$} & {16.67} & \textbf{6.67} & \textbf{55.60} & \textbf{45.00} & \textbf{30.98} \\
$\varepsilon {=} 0.7$    & 16.67 & 3.33 & 54.20 & \textbf{45.00} & 29.80 \\
$\varepsilon {=} 1$ (Dr.GRPO) & 13.33 & \textbf{6.67} & 54.40 & 40.00 & 28.60 \\
\midrule

\rowcolor{Gainsboro} \multicolumn{6}{l}{\texttt{Qwen3-4B-Base}} \\
$\varepsilon {=} 0$      & 20.00 & \textbf{23.33} & 79.40 & 72.50 & 48.81 \\
$\varepsilon {=} 0.3$    & 20.00 & 20.00 & 79.60 & 75.00 & 48.65 \\
{$\varepsilon {=} 0.5$} & \textbf{23.33} & \textbf{23.33} & \textbf{80.80} & \textbf{80.00} & \textbf{51.87} \\
$\varepsilon {=} 0.7$    & 20.00 & \textbf{23.33} & 79.20 & 75.00 & 49.38 \\
$\varepsilon {=} 1$ (Dr.GRPO) & 16.67 & \textbf{23.33} & 78.60 & 70.00 & 47.15 \\
\midrule

\rowcolor{Gainsboro} \multicolumn{6}{l}{\texttt{Qwen3-8B-Base}} \\
$\varepsilon {=} 0$      & 26.67 & 20.00 & 80.40 & 70.00 & 49.27 \\
$\varepsilon {=} 0.3$    & \textbf{33.33} & \textbf{26.67} & 83.20 & 75.00 & 54.55 \\
{$\varepsilon {=} 0.5$} & \textbf{33.33} & \textbf{26.67} & \textbf{83.20} & \textbf{77.50} & \textbf{55.17} \\
$\varepsilon {=} 0.7$    & 30.00 & \textbf{26.67} & 83.00 & 75.00 & 53.67 \\
$\varepsilon {=} 1$ (Dr.GRPO) & 26.67 & 23.33 & 81.60 & 72.50 & 51.02 \\
\bottomrule
\end{tabular}
}
\caption{Experimental results with pass@8 metric (\%) under different $\varepsilon$-exploration possibilities.}
\label{tab:epsilon_study}
\end{table}

\subsection{Impact of $\varepsilon$-exploration}\label{sec:varepsilon}

We investigate the impact of different values of $\varepsilon$ in $\varepsilon$-exploration on model performance, with the results reported in Table~\ref{tab:epsilon_study}. 
When $\varepsilon = 1$, i.e., each rollout is sampled independently, our method degenerates into the Dr.GRPO algorithm. 
We observe that, across all backbone models, performance first improves and then degrades as $\varepsilon$ increases. 
When $\varepsilon = 0$, exploration becomes overly local, which hinders diversity in the model’s exploration. 
Conversely, when $\varepsilon = 1$, the tree-structured exploration is lost, preventing effective segment-level advantage estimation.
\section{Conclusion}
In this work, we presented \MethodName, a novel reinforcement learning framework designed to enhance both the reasoning accuracy and efficiency of LLMs.
Specifically, to encourage more diverse reasoning exploration, our method incorporates a semantic-entropy-based branching strategy alongside an $\varepsilon$-exploration mechanism.
Simultaneously, to improve efficiency, we design a length-aware segment-level advantage estimator that promotes concise reasoning paths.
Extensive experiments across various mathematical benchmarks validate that ROSE significantly outperforms state-of-the-art baselines in both effectiveness and efficiency.
\section*{Limitations}
Our work mainly has two limitations.
First, our experiments were conducted on models with up to 8B parameters, and we plan to investigate the scalability on larger architectures (e.g., 14B) in future work.
Second, we primarily focused on mathematical reasoning tasks.
In the future, we plan to extend our approach to other domains, such as code generation and question answering.
\section*{Ethical Considerations}
This work aims to enhance the reasoning capabilities of LLMs.
We acknowledge that advanced reasoning abilities could potentially be misused for malicious purposes, and we advocate for the deployment of these models alongside robust safety alignment protocols to mitigate such risks.
Regarding the experimental setup, all datasets utilized in this work are open-source and publicly available.
We have strictly adhered to their respective licenses and ensured that our usage is consistent with their intended purposes.

\bibliography{custom}

\newpage
\appendix

\section{Case Study}\label{sec:case_study}
\begin{figure*}[t]
\centering
  \includegraphics[width=1\linewidth]{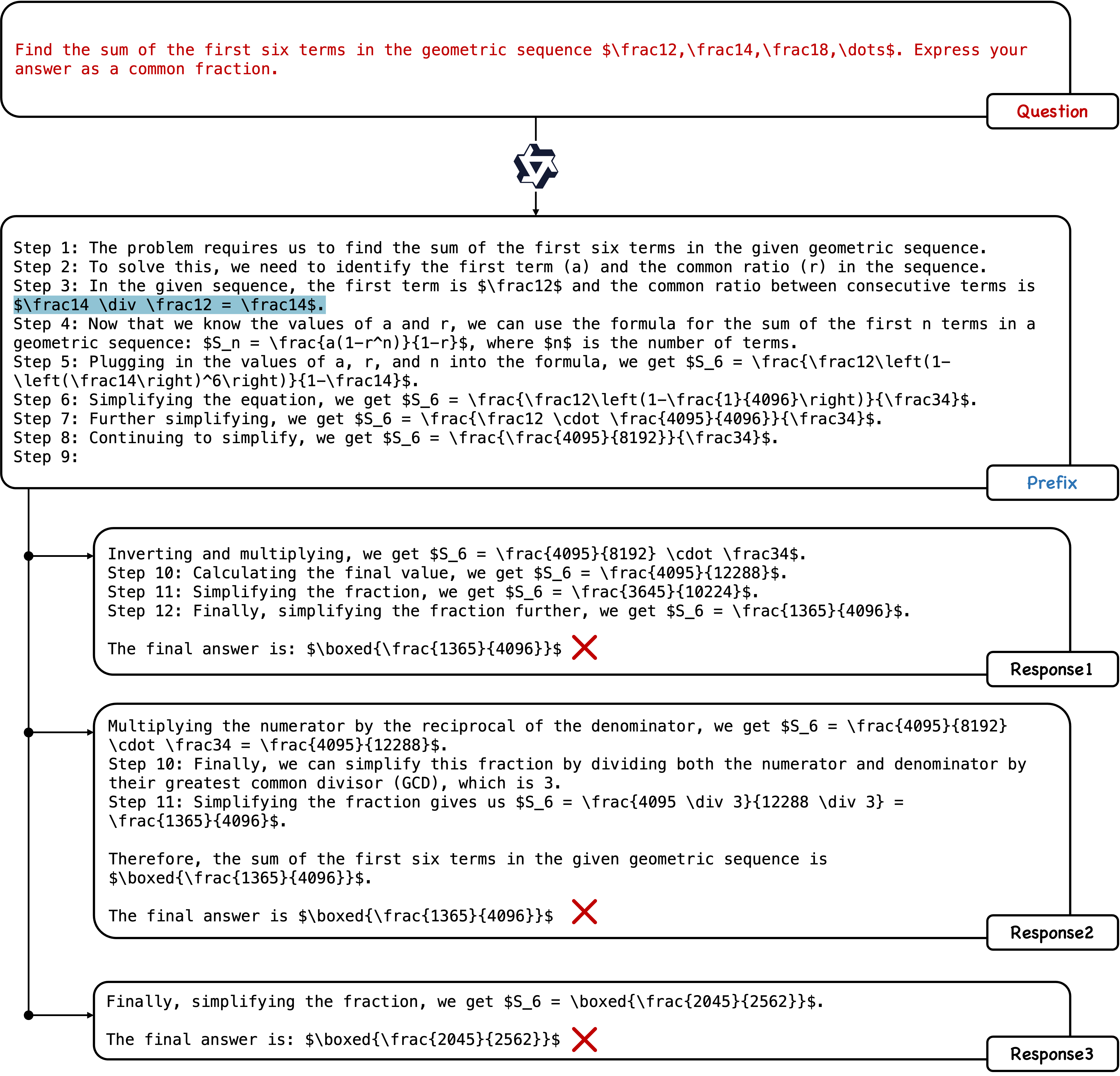}
  \caption {Case study. An example where entropy is used as the branching metric in the rollout phase.}
  \label{fig:case_entropy}
\end{figure*}

\begin{figure*}[t]
\centering
  \includegraphics[width=1\linewidth]{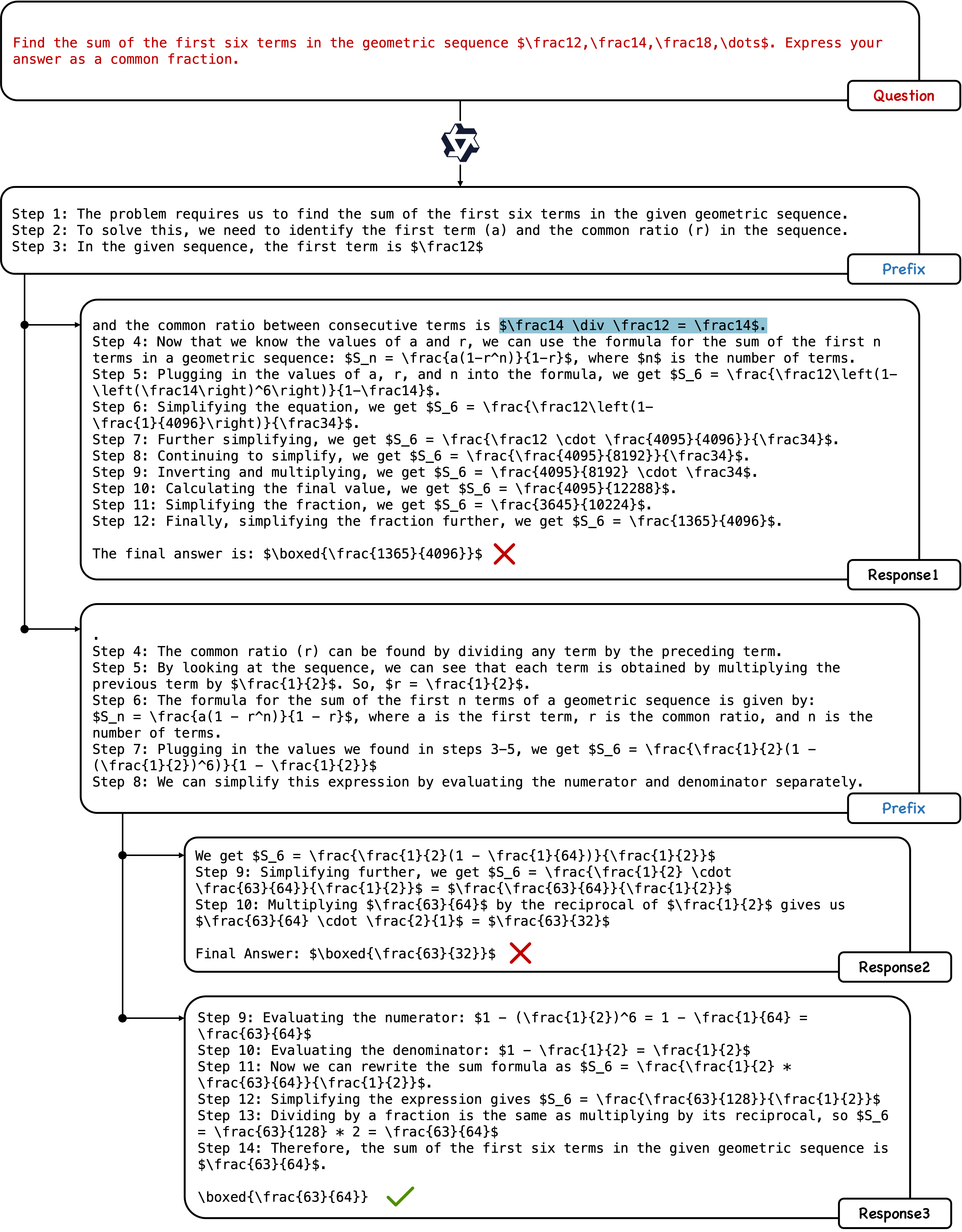}
  \caption {Case study. An example where semantic entropy is used as the branching metric in the rollout phase.}
  \label{fig:case_semantic-entropy}
\end{figure*}

To more comprehensively investigate the differences between entropy-based branching and semantic-entropy–based branching, we present a case study, with the results of the two methods shown in Figure~\ref {fig:case_entropy} and Figure~\ref{fig:case_semantic-entropy}, respectively.
Specifically, given a question, we use \texttt{Llama-3.2-3B-Instruct} as the backbone model to generate a complete response. 
We then apply two different methods to determine the branching positions based on this response and regenerate accordingly, ultimately forming a group of responses.

From Figure~\ref{fig:case_entropy}, we observe that all responses generated by the entropy-based branching method produce incorrect answers. 
Notably, the shared prefix among these responses already contains an erroneous calculation (highlighted in blue). 
However, the branching position identified by entropy-based branching occurs after this point, which consequently propagates the incorrect reasoning into the subsequent generations.

In contrast, our proposed semantic-entropy–based branching method performs branching before the erroneous calculation (highlighted in blue), enabling more fine-grained reasoning that avoids this error. 
Although subsequent reasoning errors may still occur (e.g., response 2 in Figure~\ref {fig:case_semantic-entropy}), branching is again triggered prior to the error, ultimately yielding a correct response (e.g., response 3 in Figure~\ref {fig:case_semantic-entropy}). 
This case study demonstrates that our method can more accurately identify regions of higher uncertainty in the model’s reasoning trajectory, thereby encouraging more diverse and effective reasoning paths.

\end{document}